\pdfoutput=1

\documentclass[11pt]{article}
\usepackage[final]{acl}

\usepackage{times}
\usepackage{latexsym}

\usepackage[T1]{fontenc}

\usepackage[utf8]{inputenc}

\usepackage{microtype}

\usepackage{inconsolata}

\usepackage{graphicx}
\usepackage{algorithm}
\usepackage{algorithmic}
\usepackage{amsmath}
\usepackage{amsfonts}
\usepackage{subfig}
\usepackage{booktabs}
\usepackage{multirow}
\makeatletter
\renewcommand{\maketag@@@}[1]{\hbox{\m@th\normalsize\normalfont#1}}%
\makeatother

\title{T3DM: Test-Time Training-Guided Distribution Shift Modelling for Temporal Knowledge Graph Reasoning}

\author{
 \textbf{Yuehang Si\textsuperscript{1}},
 \textbf{Zefan Zeng\textsuperscript{1}},
 \textbf{Jincai Huang\textsuperscript{1}},
 \textbf{Qing Cheng\textsuperscript{1}},
\\
\\
 \textsuperscript{1}National University of Defense Technology, Changsha, Hunan, China \\
\\
   \textbf{Correspondence:} \href{mailto:siyuehang@nudt.edu.cn}{siyuehang@nudt.edu.cn}
}

\begin{document}
\maketitle
\begin{abstract}
Temporal Knowledge Graph (TKG) is an efficient method for describing the dynamic development of facts along a timeline. Most research on TKG reasoning (TKGR) focuses on modelling the repetition of global facts and designing patterns of local historical facts. However, they face two significant challenges: inadequate modeling of the event distribution shift between training and test samples, and reliance on random entity substitution for generating negative samples, which often results in low-quality sampling. To this end, we propose a novel distributional feature modeling approach for training TKGR models, \textbf{T}est-\textbf{T}ime \textbf{T}raining-guided \textbf{D}istribution shift \textbf{M}odelling (T3DM), to adjust the model based on distribution shift and ensure the global consistency of model reasoning. In addition, we design a negative-sampling strategy to generate higher-quality negative quadruples based on adversarial training. Extensive experiments show that T3DM provides better and more robust results than the state-of-the-art baselines in most cases.
\end{abstract}

\section{Introduction}
TKGs, as a knowledge representation framework, possess the distinct capability to store and process entity-relationship information enriched with a temporal dimension. TKGs are increasingly critical in various sub-domains such as information retrieval, and recommender systems \cite{target4}. In addition, TKG's application scope extends to important areas such as policy making, dialogue systems, and stock market forecasting \cite{target1}.

\begin{figure}
	\centering
		\includegraphics[scale=.38]{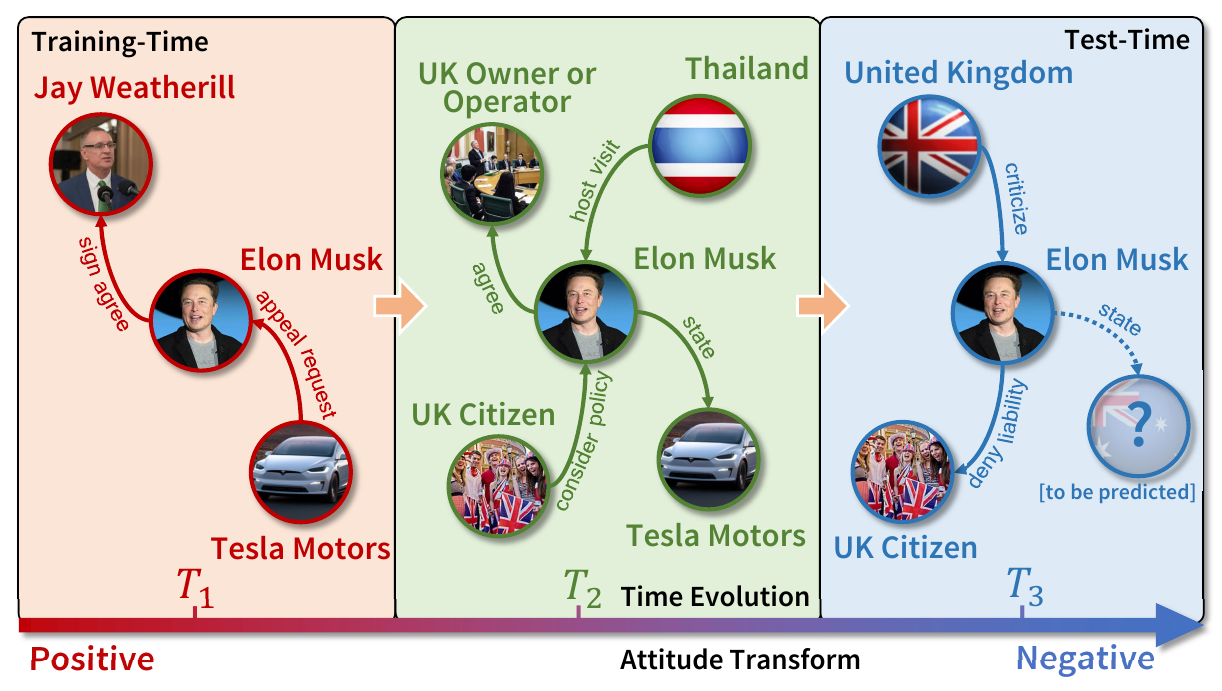}
	\caption{Illustration of the change of events in the evolution of the TKG over time in ICEWS18.}
	\label{FIG:1}
\end{figure}

The data structure of the TKG encapsulates event knowledge in the form of a quadruple $(subject,predicate,object,time)$. TKG consists of a series of snapshots of the knowledge graph, containing all events occurring simultaneously. For example, in Figure \ref{FIG:1}, ``Elon Musk'' receives an appeal request from Tesla Motors at moment $T_1$ and responds to Tesla Motors at moment $T_2$. In addition, events of ``Elon Musk'' related to UK Citizens at moments $T_2$ and $T_3$ are ``Consider policy'' and ``Deny liability'', respectively, where the roles of subject and object switch. Our study focuses on predicting future unknown events in TKG.

\begin{figure}
	\centering
		\includegraphics[scale=.38]{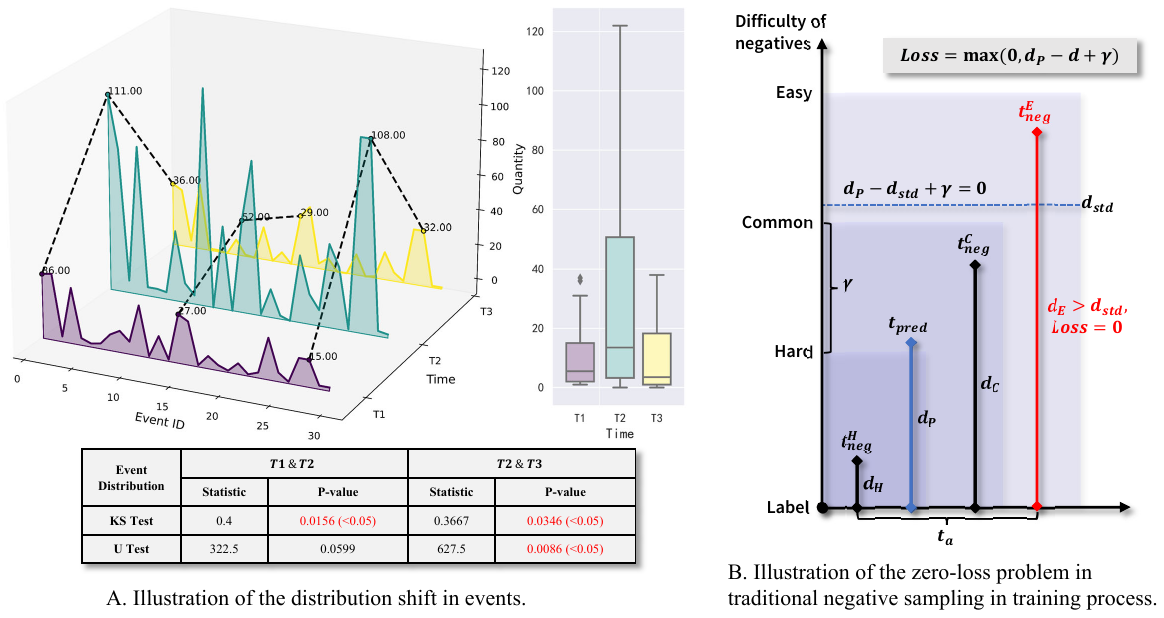}
	\caption{Illustration of event distribution shift and low negative sampling quality in ICEWS18.}
	\label{FIG:2}
\end{figure}

Recently, most existing TKGR methods encode the temporal evolution of factual relationships through time-embedded event triad data, offering a versatile and effective approach for predicting future facts on TKGs based on historical information \cite{Zhu2021}. Despite progress, TKGR field still faces two major challenges: event distribution shift and low negative sampling quality, as is shown in Figure \ref{FIG:2}.

\textbf{(1) Event Distribution Shift.} In TKG, the dynamic evolution of events is as a core research focus. Over time, the distribution of event types in TKGs shows a significant trend of change, a phenomenon known as event distribution shift. Event distribution shift refers to the change in the frequency of various types of events across distinct time periods, which can reflect the tendency of the social environment or policy tone at a specific moment. In most TKG training and test sets, event distribution shift is particularly significant. The Figure \ref{FIG:2} demonstrate the distribution shift of event relationships and results of non-parametric statistical tests (KS test and U test) for evaluating the divergence of event distributions at different time points in ICEWS18. This instance show that the event distributions exhibit shift in the time intervals from $T_1$ to $T_2$ and from $T_2$ to $T_3$. This shifting phenomenon poses a challenge to the trained TKGR model in adapting to the new event distributions under the latest time. Therefore, it is particularly crucial to empower reasoning models with the ability to adapt to shifting event distributions. Some of existing methods model the evolution of event distributions to some extent, but they do not directly consider the problem of event distribution shift \cite{Zhu2021}.

\textbf{(2) Low Negative Sampling Quality.} Most of classical TKGR models are primarily trained by maximizing the distance between negative samples and minimizing the distance between positive samples. These methods generates negative samples by replacing one of the entities with another entity in the knowledge graph while keeping the relationship and timestamp unchanged \cite{CENET}. However, the vast majority of current TKGR models rely on the random negative sampling method, i.e., by randomly selecting other entities to replace the entities in the original fact quadruple, thus obtaining negative samples. This method ignores the logical relationships and structural features among entities, thus limiting the further improvement of the model performance due to the issue of “zero-loss” shown in Figure \ref{FIG:2}.

In this study, we propose a plug-and-play training method for distribution modelling in a test-time training (TTT) framework, named T3DM. We introduce LSTM (Long Short-Term Memory) as the distributional inference model to design an auxiliary training task in the testing phase. In addition, we propose an adversarial negative-sampling strategy for TKGR, called \textbf{T}emporal \textbf{K}nowledge Graph \textbf{G}enerative \textbf{A}dversarial \textbf{N}etworks (TKGAN). We introduce a reinforcement learning algorithm to guide the sample generation process of TKGAN. We integrate T3DM into multiple TKGR baselines and conduct experiments on five publicly available datasets. The main contributions are as follows:

1) We highlight the challenges of event distribution shift and low negative sampling quality faced by existing TKGR models, which hinder their reasoning accuracy.

2) We propose a TTT-guided distribution modelling training method, T3DM, to address the event distribution shift issue in TKG. To the best of our knowledge, \textbf{we are the first to introduce the concept of TTT into knowledge graph domain to model the event distribution shift.}

3) We design TKGAN, an adversarial negative sampling strategy for TKGR, to improve the quality of negative sampling through adversarial training and incorporate a reinforcement learning strategy to guide training.

4) We conduct extensive experiments on five TKG datasets for reasoning tasks and verify the effectiveness of T3DM.

\begin{figure*}
	\centering
		\includegraphics[scale=.6]{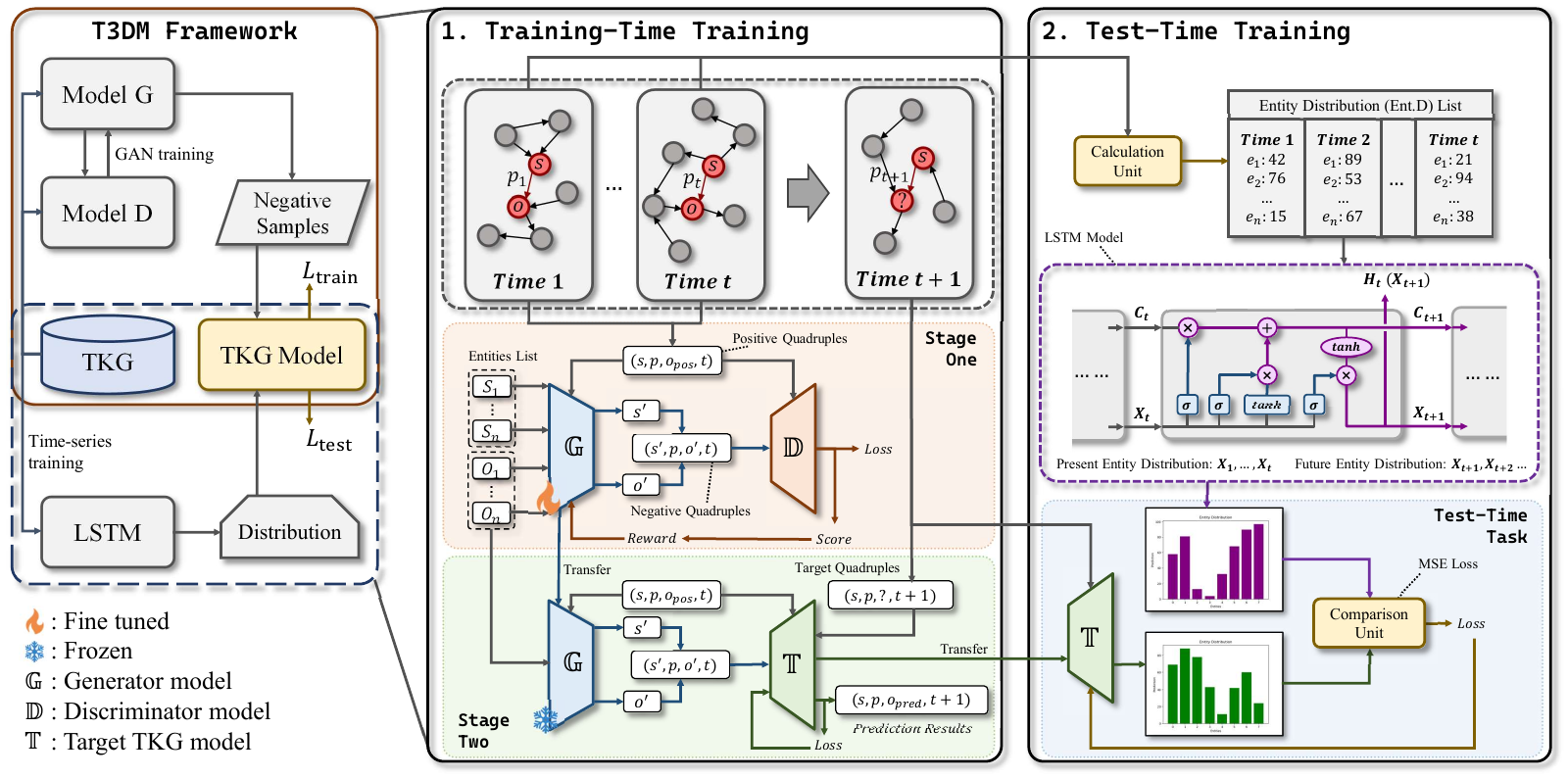}
	\caption{The training framework of T3DM. The first part represents Training Time Training, and the second part represents Test Time Training. The TKGAN part of Training Time Training is further divided into two stages.}
	\label{FIG:3}
\end{figure*}

\section{Related Work}
\subsection{Temporal Knowledge Graph Reasoning}
Temporal attributes play a crucial role in TKGR and have garnered significant attention. Know-Evolve \cite{TrivediDWS17}, as the first model to learn the nonlinear evolution of entities, lays the foundation for subsequent research. xERTE \cite{HanCMT21} and TLogic \cite{LiuMHJT22}, while providing interpretable predictive evidence, are limited in their application. TANGO \cite{HanDMGT21} utilises a neural frequent formula to model TKG, while CyGNet \cite{Zhu2021} identifies high-frequency repetitive events through the replication construction mechanism. RE-GCN \cite{LiJLGGSWC21} employs a reinforcement learning design. Some models attempt to incorporate neural network architectures to capture spatio-temporal patterns, such as RE-NET \cite{JinQJR20}, HIP \cite{HeZL0ZZ21}, TITer \cite{SunZMH021} and EvoKG \cite{ParkLMCFD22}. In addition, CENET \cite{CENET} integrates contrastive learning, while Coherence-Mode \cite{syh} designs new synergistic relationship assessment units to mine the deeper correlations. However, when the event type distribution in the test set differs significantly from that in the training set, the inference performance of these models is severely constrained.

\subsection{Test Time Training}
TTT \cite{SunWLMEH20} works for solving the problem of distribution shift between training and test samples, which is based on partially tuning the test samples to optimise the model for changes in distributions between the training and test sets \cite{Yu2019}. Meanwhile, TTT++ \cite{LiuKDBMA21} used a regularised adaptive approach of offline feature extraction and online feature alignment. TTT-MAE \cite{GandelsmanSCE22} uses a masked auto-encoder to solve the single-sample learning problem. TTT is also widely used in many domains. For example, LMTTT \cite{Zhang0Y0FSRZ024} innovatively uses the large language model as an annotator to augment the TTT. DT3OR \cite{abs15620} addresses the problem of significant degradation in recommender systems due to shifts in user and item features. Therefore, it is of significance to address distributional shifts in TKGR using TTT techniques.

\subsection{GAN for Negative Sampling}
Generative Adversarial Network is a deep learning technique proposed by Ian Goodfellow \cite{gan} in 2014. It consists of two parts: the generator and the discriminator, which improve each other's performance through adversarial training. GANs are effective in enhancing the quality of negative samples in several domains. ANDA \cite{RuizKT19} designs data-efficient GANs to improve their negative sampling ability in image classification tasks. CDE-GAN \cite{ChenWXYPCD21} uses adversarial techniques to generate high-quality negative samples in novelty detection. NDAGAN \cite{ShayestehI22} successfully applies negative data enhancement techniques to text categorisation. GAN is gradually becoming a popular negative sampling method in knowledge graph. KBGAN \cite{CaiW18} uses a knowledge graph embedding model as a negative sample generator to assist in training a reasoning model. FedEAN \cite{MengLYLZLL24} is an entity-aware negative sampling strategy through joint training of the generator and the discriminator. However, none of them takes time into account, and applies GANs to negative sampling in TKGR.

\section{The Proposed Framework}
\subsection{Overview}
The training framework of T3DM consists of two parts: training time training and test time training (see Figure \ref{FIG:3}).

Training time training is the training process of traditional TKGR based on existing quadruples to reason about future quadruples. The whole phase of the training process is structured in two stages. The first stage is composed of generator and discriminator. The generator receives the positive samples and the set of entities and generates the corresponding negative samples. The generated negative samples are used as inputs along with the positive samples for discriminator training. The discriminator's training process is the same as that of the traditional TKGR model, which maximises the weights of the positive and minimises negative samples' weights. The second stage is based on the trained generator and the target model to be trained. The target model is co-trained using the negative samples generated by the generator and the positive samples provided by the dataset with same training objective as the discriminator.

In the phase of test time training. We obtain a list of the distribution of the number of entities at different moments through the computational unit. Subsequently, LSTM is trained to predict entity number distributions at future moments. When the target model predicts the test set, the distribution of entities predicted is similarly calculated. The two distribution predictions are compared to calculate the loss, and the optimisation of the performance of the target model is continued instead.

\subsection{Preliminaries}
In TKGs, each fact contains a relation (or predicate) $p\in\mathcal{R}$ between the subject $s\in\mathcal{E}$ and the object $o\in\mathcal{E}$, which is located at time step $t\in\mathcal{T}$. Here, $\mathcal{E}$ and $\mathcal{R}$ denote the vocabularies of entities and relations, respectively, while $\mathcal{T}$ denotes the collection of timestamps. $\mathcal{G}_T$ denotes a snapshot of the TKG at time $T$, and $g=(s,p,o,t)$ denotes a quadruple (fact) in $\mathcal{G}_T$.The TKG is constructed around a set of events. These quadruple events are ordered chronologically, i.e., $\mathcal{G}=\{\mathcal{G}_1,\mathcal{G}_2, ... ,\mathcal{G}_T\}$.

The purpose of making predictions about omitted time facts is to infer that the omitted object entity $(s,p,? ,t)$ (or the supplied subject entity $(? ,p,o,t)$, or the predictive relationship $(s,? ,o,t)$. Our model is described as predicting missing entities in time facts.

\subsection{Adversarial Negative Sampling}
Inspired by GAN, we propose a TKG adversarial training framework, TKGAN, for adversarial generation of negative quadratic samples. Corresponding to the terminology used in the classical GAN literature, in the rest of this paper, we refer to the two models used for adversarial training simply as the generator and discriminator, respectively. Our work has a similar training goal as the classical GAN framework: the ultimate goal of our framework is to train a good generator to generate high-quality negative samples. The discriminator should assign a relatively high score to high-quality negative samples during training. Therefore, the goal of the generator should be set to maximise the score given by the discriminator for the quadruples it generates. Just like the traditional training process for TKGR models, the discriminator aims to minimise the marginal loss between the positive quadruple and the generated negative quadruple. In an adversarial training environment, the generator and discriminator are alternately trained on their respective goals.

Suppose that the generator generates a probability distribution $p_{\mathbb{G}}(s^{\prime},p,o^{\prime},t\mid s,p,o,t)$ over the negative quadruple $(s^{\prime},p,o^{\prime},t)$ given the positive quadruple $(s,p,o,t)$, and by sampling from this distribution to generate the negative quadruple $(s^{\prime},p,o^{\prime},t)$. Let $f_{\mathbb{D}}(s,p,o,t)$ be the score function of the discriminator. The objective of discriminator can be stated as minimising the following marginal loss function:
\begin{equation}
\begin{split}
L_{\mathbb{D}}=\sum_{g\in \mathcal{G}} \left[f_{\mathbb{D}}(g)-f_{\mathbb{D}}\left(s^{\prime},p,o^{\prime},t\right)+\gamma\right]_{+},
\end{split}
\end{equation}
where $(s^{\prime}, p, o^{\prime},t)\sim p_{\mathbb{G}}(s^{\prime},p,o^{\prime},t\mid s,p,o,t )$ is the negative quadruple generated by generator, $g=(s,p,o,t)$, $\gamma$ is the constant offset coefficient, and $[.] _+$ denotes the take positive operation.

The computation of the reward of the generator can be formulated as maximising the expectation of the negative distance of the discriminator:
\begin{equation}
R_{\mathbb{G}}=\sum_{(s, p, o,t)\in \mathcal{G}} E\left(-f_\mathbb{D}\left(s^{\prime}, p, o^{\prime},t\right)\right),
\end{equation}
where $(s^{\prime}, p, o^{\prime},t)\sim p_{\mathbb{G}}(s^{\prime},p,o^{\prime},t\mid g)$, $E(.) $ denotes the expectation operation.

The $R_\mathbb{G}$ involves a discrete sampling step, and the gradient cannot be computed by simple differentiation. We use the strategic gradient theorem to obtain the gradient of $R_\mathbb{G}$ with respect to the generator parameters:
\begin{equation}
\begin{split}
&\nabla_\mathbb{G} R_\mathbb{G}=
\sum_{(s, p,o, t)\in \mathcal{G}} \mathbb{E}_{\left(s^{\prime}, p, o^{\prime},t\right)\sim p_\mathbb{G}\left(s^{\prime}, p, o^{\prime},t\mid g\right)}\\
&{\left(-f_\mathbb{D}\left(s^{\prime}, p, o^{\prime},t\right)\nabla_\mathbb{G}\log p_\mathbb{G}\left(s^{\prime}, p, o^{\prime},t\mid g\right)\right)}\\& 
\simeq\sum_{g\in \mathcal{G}}\frac{1}{N}\sum_{\left(s_i^{\prime}, p, o_i^{\prime},t\right)\sim p_\mathbb{G}\left(s^{\prime}, p, o^{\prime},t\mid g\right), i=1\ldots N}
\\&{\left(-f_\mathbb{D}\left(s^{\prime}, p, o^{\prime},t\right)\nabla_\mathbb{G}\log p_\mathbb{G}\left(s^{\prime}, p, o^{\prime},t\mid g\right)\right)}.
\end{split}
\end{equation}

The approximation equation shows that we can approximate the expectation with sampled mean.  This approximation implements the gradient computation of $R_\mathbb{G}$ and enables its optimisation using gradient-based algorithms.

The strategy gradient theorem is derived from reinforcement learning (RL). Thus, the generator can be viewed as an agent that interacts with environment by performing actions and improves itself by maximising the reward for its actions. Correspondingly, the discriminator serves as the environment, and its output corresponds to the feedback made by the environment. Using RL terminology, $g=(s,p,o,t)$ is the state (determining what actions the agent can take), $p_\mathbb{G}(s^{\prime},p,o^{\prime},t\mid g)$ is the strategy (how the agent chooses to take action), and $(s^{\prime},p,o^{\prime},t)$ is the action and $-f_\mathbb{D}(s^{\prime},p,o^{\prime},t)$ is the reward. Unlike typical RL, where agent performs a series of actions, the agent in our model acts only once and does not affect the state.

To reduce the variance of the gradient algorithm, a baseline value, which depends solely on the state of the agent, is typically subtracted from its reward. In our model, we introduce a baseline term in $-f_D(s^{\prime},p,o^{\prime},t)$. We replace it with $-f_D(s^{\prime},p,o^{\prime},t)-b(g)$. $b$ is a constant, the reward over the entire training set, approximated by the average of the rewards of the most recently generated negative quadruples:
\begin{equation}
\begin{split}
b=\sum_{g \in \mathcal{G}} \mathbb{E}_{\left(s^{\prime}, p, o^{\prime},t\right) \sim p_\mathbb{G}\left(s^{\prime}, p, o^{\prime},t \mid g\right)}\left(-f_\mathbb{D}\left(s^{\prime}, p, o^{\prime},t\right)\right).
\end{split}
\end{equation}

Given a set of candidate negative quadruples $\operatorname{Neg}(g)\subseteq\{(s^{\prime},p,o,t)\mid s^{\prime}\in \mathcal{E}\}\cup\{(s,p,o^{\prime},t)\mid o^{\prime}\in \mathcal{E}\}$, then the probability distribution $p_G$ is modelled as:
\begin{equation}
\begin{split}
p_\mathbb{G}\left(s^{\prime}, p, o^{\prime},t\mid g\right)=\frac{\exp f_\mathbb{G}\left(s^{\prime}, p, o^{\prime},t\right)}{\sum_{s^{*},o^{*}}\exp f_\mathbb{G}\left(s^{*}, p, o^{*},t\right)}.
\end{split}
\end{equation}
where $\left(s^{*}, p, o^{*},t\right)\in\operatorname{Neg}(g)$, and $f_\mathbb{G}(g)$ is the generator score function.

\begin{algorithm}[h]
\footnotesize
\caption{The TKGAN algorithm}
\renewcommand{\algorithmicrequire}{\textbf{Input:}}
\renewcommand{\algorithmicensure}{\textbf{Output:}}
\begin{algorithmic}[1]
\REQUIRE Training set of positive fact quadruples $\mathcal{G}=\{g:(s,p,o,t)\}$; Pre-trained generator $\mathbb{G}$ and $\mathbb{D}$ with parameters $\theta_\mathbb{G}, \theta_\mathbb{D}$ and score function $f_\mathbb{G}(g), f_\mathbb{D}(g)$
\ENSURE Adversarially trained generator $\mathbb{G}$
\STATE $b \leftarrow 0$; \%baseline constant for policy gradient
\REPEAT
\STATE Sample a small batch of positive quadruples $\mathcal{G}_{\text{batch}}$ from $\mathcal{G}$
\STATE Initial gradients of parameters: $G_\mathbb{G} \leftarrow 0, G_\mathbb{D} \leftarrow 0$
\STATE Total reward: $r_{\text{total}} \leftarrow 0$
\FOR{each $g \in \mathcal{G}_{\text{batch}}$}
\STATE Uniformly randomly sample $K$ negative quadruples $\text{Neg}(g)=\{(s_{i}^{\prime},p,o_{i}^{\prime},t)\}_{i=1...K}$
\STATE Obtain probability of being generated: $p_{i}=\frac{\exp f_\mathbb{G}(s_{i}^{\prime},p,o_{i}^{\prime},t)}{\sum_{j=1}^{K}\exp f_\mathbb{G}(s_{j}^{\prime},p,o_{j}^{\prime},t)}$
\STATE Sample negative quadruple $(s_{h}^{\prime},p,o_{h}^{\prime},t)$ from $\text{Neg}(g)$ with the highest probability $p_{h}$ according to $\{p_{i}\}_{i=1...K}$
\STATE $G_\mathbb{D} \leftarrow G_\mathbb{D} + \nabla_{\theta_\mathbb{D}}[f_\mathbb{D}(g)-f_\mathbb{D}(s_{h}^{\prime},p,o_{h}^{\prime},t)+\gamma]_{+}$
\STATE Calculate reward for $\mathbb{G}$: $r \leftarrow -f_\mathbb{D}(s_{h}^{\prime},p,o_{h}^{\prime},t)$, $r_{\text{total}} \leftarrow r_{\text{total}} + r$
\STATE $G_\mathbb{G} \leftarrow G_\mathbb{G} + (r-b)\nabla_{\theta_\mathbb{G}}\log p_{h}$
\ENDFOR
\STATE $\theta_\mathbb{G} \leftarrow \theta_\mathbb{G} + \eta_\mathbb{G}G_\mathbb{G}, \theta_\mathbb{D} \leftarrow \theta_\mathbb{D} - \eta_\mathbb{D}G_\mathbb{D}$
\STATE $b \leftarrow {r_{\text{total}}}/{|\mathcal{G}_{\text{batch}}|}$
\UNTIL{convergence}
\end{algorithmic}
\label{alg1}
\end{algorithm}

Ideally, $\operatorname{Neg}(g)$ should contain all possible negative quadruples. However, TKGs are usually highly incomplete, so the ``hardest to distinguish'' negative quadruples are likely to be false negatives (true facts). To address this problem, we generate $\operatorname{Neg}(g)$ by uniformly sampling a certain number of entities (a minimal number compared to the number of all possible negatives) from the entity set $\mathcal{E}$ to replace either $s$ or $t$. As in real-world TKGs, the quantity of true negative quadruples significantly exceeds that of false negative quadruples. Consequently, it is improbable that a given set includes any false negative samples. Furthermore, the negative quadruples selected by the generator are highly likely to be classified as true negative quadruples.

In addition, we employ the ``bern'' sampling technique \cite{hyperplanes} to reduce false negatives further by replacing ``1'' of relationships with higher probability in ``1-to-N'' and ``N-to-1''. Algorithm 
\ref{alg1} details the steps of TKGAN.

In the training phase, T3DM uses the negative quadruple samples generated based on the TKGAN method as a new part of generating negative samples in the traditional TKGR. Note that the choice of generator and discriminator for TKGAN is very flexible, and the target model to be trained can either be directly used as a discriminator or co-trained with the generator model. In our approach T3DM, we separate the training process of the target model to be trained from the generator model, i.e., the negative sample generator model is trained first (Stage 1) and then added to the training of the target model (Stage 2). This design allows T3DM to adapt more effectively by directly generating negative samples for TKGR models, eliminating the need to adjust the corresponding relationships within TKGAN.

\subsection{Test-Time Training}
To address the issue of changing event type distributions over time in TKGs, we integrate TTT with TKGR models. In the field of TKGR, we explore an auxiliary training strategy conducted at test-time, i.e., to improve the inference ability of the model by training its ability to predict the event type distribution during the Test-time phase. Specifically, since individual events refer to the associations generated between entities, the distribution of event types can be determined by the distribution of the number of various types of entities at a particular point in time, and the distribution of the number of entities can adequately reflect the changes in event types. This correspondence is crucial for understanding and predicting dynamic changes in TKG.

To achieve this goal, we employ LSTM to predict the distribution of  the entities as the "pseudo labels" for TKGR in the Test-time phase. LSTM consists of forget gate, input gate, cell state, and output gate, as is shown in Figure \ref{FIG:3}.
\begin{equation}
\begin{split}
  S_{t+1}=\text{LSTM}(S_{t-l+1},...,S_{t-1},S_{t})
\end{split}
\end{equation}
where $S_{t}$ denotes the event distribution at time $t$. $l$ is the input sequence length of LSTM.

\begin{table*}[h]
\scriptsize
\setlength{\tabcolsep}{0.9pt}
    \renewcommand{\arraystretch}{0.95}
  \caption{Experimental results of T3DM for TKG link prediction task on five datasets. The best results are presented in boldface, and the previous SOTA are underlined (if needed). H@1, H@3 and H@10 represent Hits@1, Hits@3 and Hits@10, respectively.}
\scriptsize
  \begin{tabular}{l|l|llll|llll|llll|llll|llll}
    \toprule
    \multirow{2}{*}{Baseline} & \multirow{2}{*}{Model} & \multicolumn{4}{l|}{ICEWS18}& \multicolumn{4}{l|}{ICEWS14}& \multicolumn{4}{l|}{GDELT}& \multicolumn{4}{l|}{WIKI}& \multicolumn{4}{l}{YAGO}\\
    \cmidrule{3-22}
    ~ & ~ & MRR & H@1& H@3 & H@10& MRR & H@1& H@3 & H@10& MRR & H@1& H@3 & H@10& MRR & H@1& H@3 & H@10& MRR & H@1& H@3 & H@10\\
    \midrule
    \multirow{11}{*}{TransE} &TTransE &8.36&1.94&8.71&21.93&6.35&1.23&5.80&16.65&5.52&0.47&5.01&15.27&31.74&32.61&36.25&43.45&32.57&\textbf{34.29}&43.39&53.37\\
    ~ &TTransE\&TKGAN &{11.92}&{3.40}&{13.82}&{29.80}&{10.24}&{2.76}&{11.28}&{26.12}&{6.78}&{0.53}&{6.62}&{20.11}&{38.41}&{37.60}&{42.85}&{47.57}&{41.84}&27.05&{54.47}&{62.65}\\
    
    ~ &TTransE\&T3DM &\textbf{11.98}&\textbf{3.50}&\textbf{13.91}&\textbf{29.87}&\textbf{10.29}&\textbf{2.83}&\textbf{11.30}&\textbf{26.13}&\textbf{6.84}&\textbf{0.58}&\textbf{6.71}&\textbf{20.30}&\textbf{38.44}&\textbf{37.65}&\textbf{42.86}&\textbf{47.59}&\textbf{42.31}&{27.66}&\textbf{54.92}&\textbf{62.77}\\
    \cmidrule{2-22}
    ~&TATransE &9.28&3.89&9.34&18.20&8.39&4.22&{9.06}&20.81&11.37&7.44&11.83&21.32&41.89&37.86&45.20&49.74&51.35&45.86&56.73&62.11\\
    ~ &TATransE\&TKGAN &{10.46}&{5.92}&{10.99}&{19.42}&{13.54}&{7.81}&{14.52}&{24.66}&{14.77}&{9.32}&{15.39}&{24.89}&{45.28}&{40.92}&{48.87}&{51.92}&{54.37}&{47.92}&{59.31}&{65.27}\\
    ~ &TATransE\&T3DM &\textbf{10.50}&\textbf{5.94}&\textbf{11.06}&\textbf{19.48}&\textbf{13.59}&\textbf{7.82}&\textbf{14.56}&\textbf{24.69}&\textbf{14.83}&\textbf{9.40}&\textbf{15.43}&\textbf{24.92}&\textbf{45.33}&\textbf{40.99}&\textbf{48.96}&\textbf{51.99}&\textbf{54.42}&\textbf{48.01}&\textbf{59.51}&\textbf{65.32}\\
    \cmidrule{2-22}
    ~&HyTE&7.31&4.03&7.50&14.95&\textbf{11.48}&4.30&13.04&22.51&6.37&4.78&6.72&18.63&43.02&27.99&45.12&49.49&23.16&36.62&45.74&51.94\\
    ~ &HyTE\&TKGAN &{8.31}&{5.00}&{9.14}&{14.96}&10.98&{4.48}&{13.17}&{23.06}&{10.11}&{6.66}&{10.88}&{18.93}&{44.56}&{28.84}&{45.87}&{50.45}&{42.86}&{38.51}&{46.47}&{52.39}\\
    ~ &HyTE\&T3DM &\textbf{8.33}&\textbf{5.04}&\textbf{9.19}&\textbf{15.00}&{11.03}&\textbf{4.50}&\textbf{13.21}&\textbf{23.14}&\textbf{10.16}&\textbf{6.70}&\textbf{10.91}&\textbf{18.94}&\textbf{44.60}&\textbf{28.87}&\textbf{45.89}&\textbf{50.46}&\textbf{42.89}&\textbf{38.60}&\textbf{46.54}&\textbf{52.40}\\
    \midrule
    \multirow{3}{*}{Distmult} &TADistmult &\textbf{28.53}&{20.30}&31.57&44.96&20.78&13.43&22.80&35.26&{29.35}&22.11&31.56&41.39&48.09&45.84&49.51&51.70&61.72&\textbf{62.80}&\textbf{65.32}&67.19\\
    ~&TADistmult\&TKGAN &27.79&20.17&{31.81}&{45.52}&{22.52}&{15.98}&{23.50}&{35.75}&29.42&{22.61}&{32.36}&{43.17}&{51.13}&{49.15}&{52.23}&{57.09}&{62.82}&59.37&63.67&{71.46}\\
    ~ &TADistmult\&T3DM &{27.93}&\textbf{20.36}&\textbf{31.86}&\textbf{45.59}&\textbf{22.60}&\textbf{16.04}&\textbf{23.55}&\textbf{35.81}&\textbf{29.46}&\textbf{22.71}&\textbf{32.48}&\textbf{43.21}&\textbf{51.19}&\textbf{49.22}&\textbf{52.36}&\textbf{57.20}&\textbf{62.99}&{59.74}&{63.81}&\textbf{71.60}\\
    \midrule
    \multirow{3}{*}{SimplE} &DE-SimplE &12.56&9.84&19.87&27.07&16.08&9.71&13.90&16.55&20.49&6.33&17.00&23.97&18.17&9.92&18.79&32.44&30.67&34.24&38.5&47.02\\
    ~&DE-SimplE\&TKGAN &{12.93}&{10.15}&{20.32}&{27.22}&{16.34}&{10.03}&{14.17}&{16.70}&{20.75}&{6.68}&{17.14}&{24.14}&{18.21}&{10.08}&{18.98}&{32.56}&{30.80}&{34.39}&{38.69}&{47.15}\\
    ~&DE-SimplE\&T3DM &\textbf{13.06}&\textbf{10.24}&\textbf{20.46}&\textbf{27.32}&\textbf{16.51}&\textbf{10.22}&\textbf{14.29}&\textbf{16.83}&\textbf{20.97}&\textbf{6.85}&\textbf{17.38}&\textbf{24.40}&\textbf{18.49}&\textbf{10.33}&\textbf{19.17}&\textbf{32.74}&\textbf{30.86}&\textbf{34.41}&\textbf{38.73}&\textbf{47.27}\\
\bottomrule
\end{tabular}
	\label{tab2}
\end{table*}

When testing, the TKGR model is used to predict the distribution of the number of entities. The prediction distribution is compared with the distribution predicted by LSTM model, and the loss between the two is calculated as follows:
\begin{equation}
\begin{split}
 &L_{cmp}=\sum_{i=1}^{T_{Pred}} {CE\_Loss (X_{i}^{True},X_{i}^{Pred})}
 \\&=-\sum_{i=1}^{T_{Pred}}\sum_{j=1}^{N}{x_{i,j}^{True}}\text{log}({\text{softmax}(P_{i,j}}),
\end{split}
\end{equation}
where $CE\_Loss(.) $ denotes the cross-entropy loss function, $N$ denotes the number of object entities in the snapshot $\mathcal{G}_t$, and $T_{Pred}$ denotes the predicting time period. $X^{True}$ denotes the distribution predicted by the LSTM, and $X^{Pred}$ denotes the distribution predicted by TKGR model. $P_{i,j}$ denotes the probability of $x_{i,j}^{Pred}$ on each entity.

By auxiliary training, the TKGR model can continuously adjust and optimise its parameters during the testing phase to improve prediction accuracy. This approach enhances the model's ability to adapt to dynamic changes in the knowledge graph and its robustness in complex temporal reasoning tasks.

\subsection{Inference}
We describe the reasoning process as predicting missing objects in time facts without loss of generality. In predicting the query $(s,p,?,t)$, the TKGR model provides the object entity with the highest probability in the candidate space. The prediction result of the original baseline model is defined as follows:
\begin{equation}
  o_t = \text{argmax}_{o_{pred}\in \mathcal{E}} P(o_{pred}\mid s,p,o_{rand},t)
\end{equation}
where $o_{pred}$ denotes the final prediction of the model and $o_{rand}$ denotes the random entity input to the model.

\section{Experiment}
This section shows the effectiveness of the proposed T3DM\footnote{The released source code and documentation are available at \hyperlink{https://anonymous.4open.science/r/T3DM-6514}{https://anonymous.4open.science/r/T3DM-6514}} through a comprehensive experiment by solving the following questions: RQ1: What is the gain of T3DM's performance on TKGR tasks? RQ2: What are the advantages of the adversarial negative sampling method used in T3DM? RQ3: Can the unique training framework based on TTT in T3DM be applied to a more extensive range of baseline models? RQ4: How can different choices of generator and discriminators in TKGAN affect model performance? RQ5: How does the prediction of LSTM models with different input sequence length affect model performance?

\subsection{Experimental Setup}
\subsubsection{Datasets and Metrics}
We select five publicly available benchmark datasets for TKGR: GDELT \cite{gdelt}, ICEWS14 \cite{icews}, ICEWS18, YAGO11K \cite{yago}, and Wikidata12K \cite{wiki}.

To evaluate the performance of the T3DM model, we use standard metrics such as Mean Reciprocal Rank (MRR) and Hits@K (K=1,3,10). MRR denotes the average inverse ranking of all samples, while Hits@K measures the proportion of test samples ranked in the top K positions.

\begin{table*}[h]
\scriptsize
\setlength{\tabcolsep}{1.8pt}
    \renewcommand{\arraystretch}{0.95}
  \caption{Ablation experimental results of TTT part in T3DM for TKG link prediction task on five datasets.}
\scriptsize
  \begin{tabular}{l|llll|llll|llll|llll|llll}
    \toprule
    \multirow{2}{*}{Model} & \multicolumn{4}{l}{ICEWS18}& \multicolumn{4}{|l}{ICEWS14}& \multicolumn{4}{|l}{GDELT}& \multicolumn{4}{|l}{WIKI}& \multicolumn{4}{|l}{YAGO}\\
    \cmidrule{2-21}
    ~ & MRR & H@1& H@3 & H@10& MRR & H@1& H@3 & H@10& MRR & H@1& H@3 & H@10& MRR & H@1& H@3 & H@10& MRR & H@1& H@3 & H@10\\
    \midrule
    TransE &17.56&2.48&26.95&43.87&18.65&1.12&31.34&47.07&16.05&0.00&26.10&42.29&46.68&36.19&49.71&51.71&48.97&46.23&62.45&66.05\\
    Distmult &22.16&12.13&26.00&42.18&19.06&10.09&22.00&36.41&18.71&11.59&20.05&32.55&46.12&37.24&49.81&51.38&49.47&52.97&60.91&65.26\\
    ComplEx &30.09&21.88&34.15&45.96&24.47&16.13&27.49&41.09&22.77&15.77&24.05&36.33&47.84&38.15&50.08&51.39&61.29&54.88&62.28&66.82\\
    RotatE &23.10&14.33&27.61&38.72&29.56&22.14&32.92&42.68&22.33&16.68&23.89&32.29&50.67&39.73&50.71&50.88&65.09&55.69&65.67&66.16\\
    \midrule
    RE-NET &42.93&36.19&45.47&55.80&45.71&38.42&49.06&59.12&40.12&32.43&43.40&53.80&51.97&48.01&52.07&53.91&65.16&63.29&65.63&68.08\\
    TANGO-TuckER &44.56&37.87&47.46&57.06&46.42&38.94&50.25&59.80&38.00&28.02&43.91&53.70&53.28&52.21&53.61&62.72&67.21&65.56&67.59&77.23\\
    TANGO-Distmult &44.00&38.64&45.78&54.27&46.68&41.20&48.64&57.05&41.16&35.11&43.02&52.58&54.05&51.52&53.84&62.95&68.34&67.05&68.39&78.10\\
    CyGNet &46.69&40.58&49.82&57.14&48.63&41.77&52.50&60.29&50.29&44.53&54.69&60.99&45.50&50.48&50.79&52.80&63.47&64.26&65.71&68.95\\
    EvoKG &29.67&12.92&33.08&58.32&18.30&6.30&19.43&39.37&11.29&2.93&10.84&25.44&50.66&12.21&63.84&67.29&55.11&54.37&81.38&83.81\\
    CENET &51.06&47.10&51.92&58.82&53.35&\underline{49.61}&54.07&60.62&58.48&55.99&58.63&\textbf{62.96}&68.39&68.33&68.36&69.05&\underline{84.13}&84.03&\underline{84.23}&85.47\\
    HIP network&48.37&43.51&51.32&58.49&50.57&45.73&54.28&\textbf{61.65}&52.76&46.35&55.31&61.87&54.71&53.82&54.73&56.46&67.55&66.32&68.49&70.37\\
    CEC-BD &28.53&18.85&32.31&47.75&47.53&39.77&53.25&59.54&34.74&27.25&39.37&52.21&33.93&24.12&36.92&54.33&21.26&15.44&21.58&33.99\\
    Co-CyGNet&47.93&43.21&51.38&58.99&49.57&43.04&53.64&60.03&50.77&45.05&54.93&61.42&46.31&52.19&52.34&54.11&63.83&65.14&66.49&70.03\\
    Co-CENET&51.47&\underline{47.32}&52.08&\underline{59.11}&\underline{53.37}&49.58&\underline{54.81}&61.30&\underline{58.63}&56.31&58.99&62.88&68.77&68.41&68.58&69.34&84.08&\underline{84.25}&84.19&\underline{85.56}\\
    \midrule
    CyGNet\&TTT&47.85&43.36&51.16&58.04&49.30&43.28&53.33&60.31&50.91&45.12&54.99&60.96&46.73&52.16&52.53&54.16&67.66&66.06&68.23&70.01\\
    CENET\&TTT&\underline{51.49}&47.31&\underline{52.10}&59.01&53.35&49.55&\underline{54.81}&61.33&58.61&\underline{56.35}&\underline{59.00}&62.79&\underline{68.78}&\underline{68.44}&\underline{68.60}&\underline{69.39}&84.06&84.23&84.17&85.55\\
    Co-CENET\&TTT&\textbf{51.51}&\textbf{47.34}&\textbf{52.12}&\textbf{59.15}&\textbf{53.41}&\textbf{49.64}&\textbf{54.83}&\underline{61.38}&\textbf{58.65}&\textbf{56.37}&\textbf{59.02}&\underline{62.91}&\textbf{68.80}&\textbf{68.49}&\textbf{68.64}&\textbf{69.40}&\textbf{84.19}&\textbf{84.27}&\textbf{84.25}&\textbf{85.58}\\
\bottomrule
\end{tabular}
	\label{tab3}
\end{table*}

\subsubsection{Baselines}
To validate the effectiveness of T3DM, we choose a series of baseline models to start experiments, including TTransE \cite{TTransE}, TATransE \cite{TADistmult}, HyTE \cite{HyTE}, TADistmult \cite{TADistmult} and DE-SimplE \cite{DESimplE}. We also compare with a range of static KG and TKGR models, including TransE \cite{TransE}, DistMult \cite{Distmult}, ComplEx \cite{ComplEx}, RotatE \cite{RotatE}, RE-NET \cite{JinQJR20}, TANGO \cite{HanDMGT21}, CyGNet \cite{Zhu2021}, EvoKG \cite{ParkLMCFD22}, CENET \cite{CENET}, HIP Network \cite{HeZL0ZZ21}, CEC-BD \cite{YueRZZZWZ24}, and Coherence-Mode \cite{syh}.

\subsection{Experimental Performance (RQ1)}
To address RQ1, we analyse the performance of the proposed models. Specifically, we integrate the proposed T3DM into three types of baselines totalling five models. We compare the performance of these baselines on five TKG datasets with the integration effect. Table \ref{tab2} shows the results of using TTransE as the generator. It can be seen that the integration of T3DM obtains entire performance gains on both TATransE and DE-SimplE (achieve performance improvements of 3.04 and 0.37 on average respectively), and only one metric is slightly inferior to the baseline on TTransE and HyTE. These results suggest that the TKG model can significantly improve its ability to reason about future events by incorporating the antagonistic negative sampling approach and adding training on factual distributions at test time. In contrast, a small number of metrics are down on TADistmult (with a lag on YAGO, from 62.80 to 59.74 on H@1), which is mainly attributed to the fact that the architecture of TADistmult and its handling of negative samples is very different from TTransE, and the return rewards have limited enhancement on negative sampling. Nevertheless, in terms of the experiment as a whole, baselines integrated with T3DM still show a significant performance improvement.

\subsection{Ablation Study (RQ2 \& RQ3)}
\begin{figure}
    \centering
    \subfloat{\includegraphics[width=0.12\textwidth]{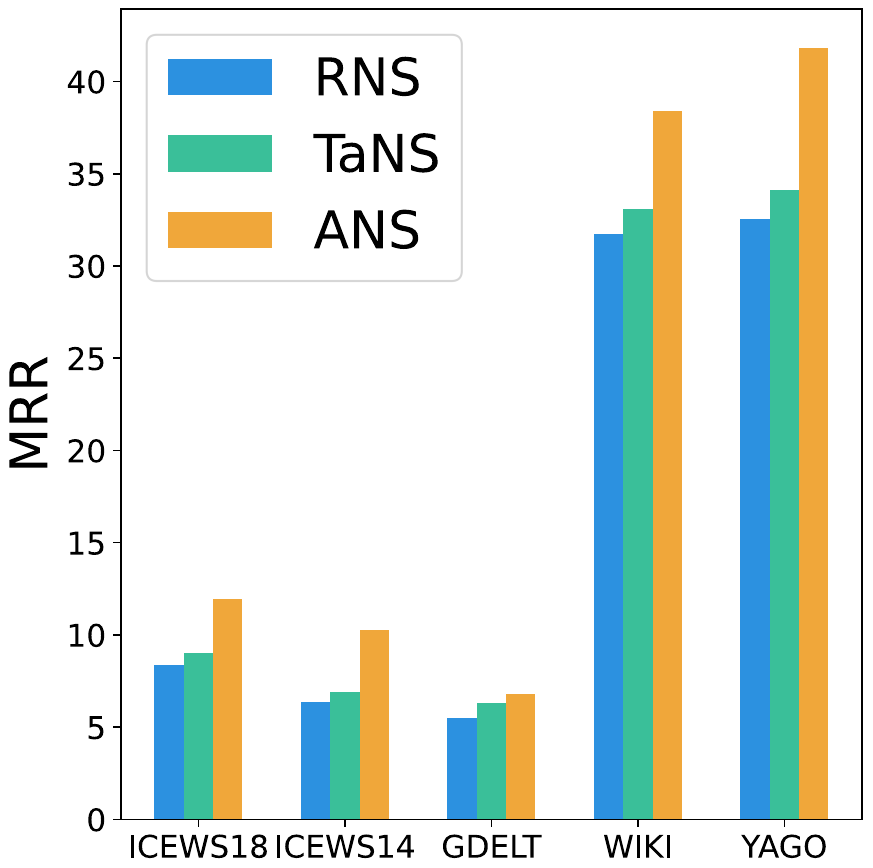}}
    \hfill
    \subfloat{\includegraphics[width=0.12\textwidth]{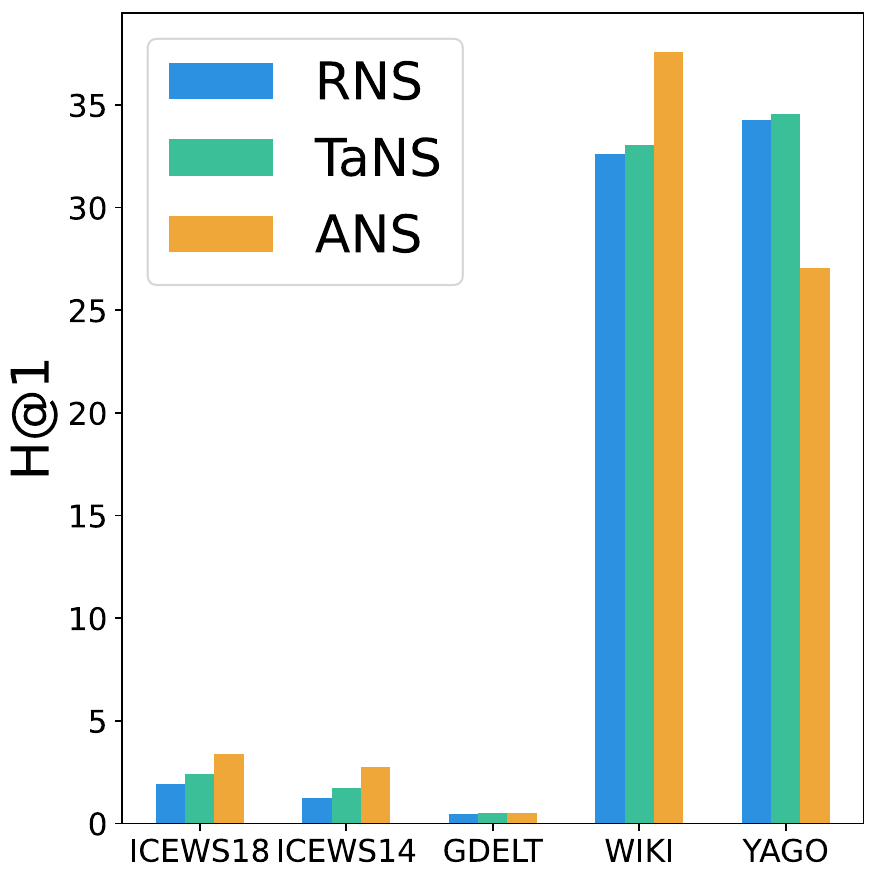}}
    \hfill
    \subfloat{\includegraphics[width=0.12\textwidth]{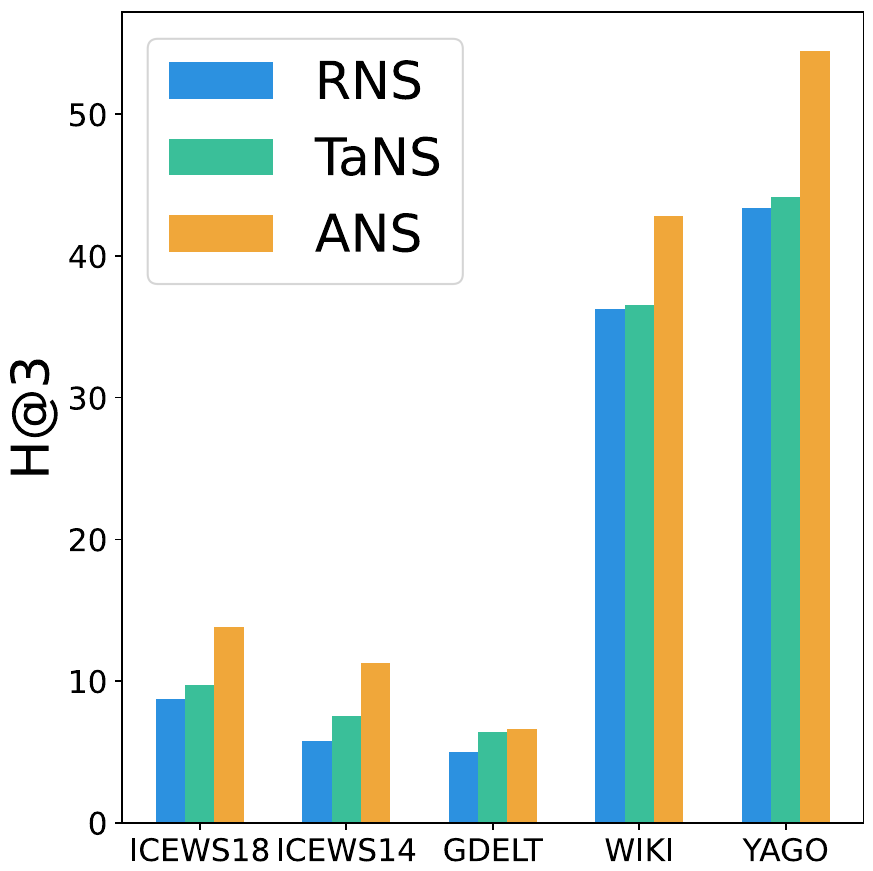}}
    \hfill
    \subfloat{\includegraphics[width=0.12\textwidth]{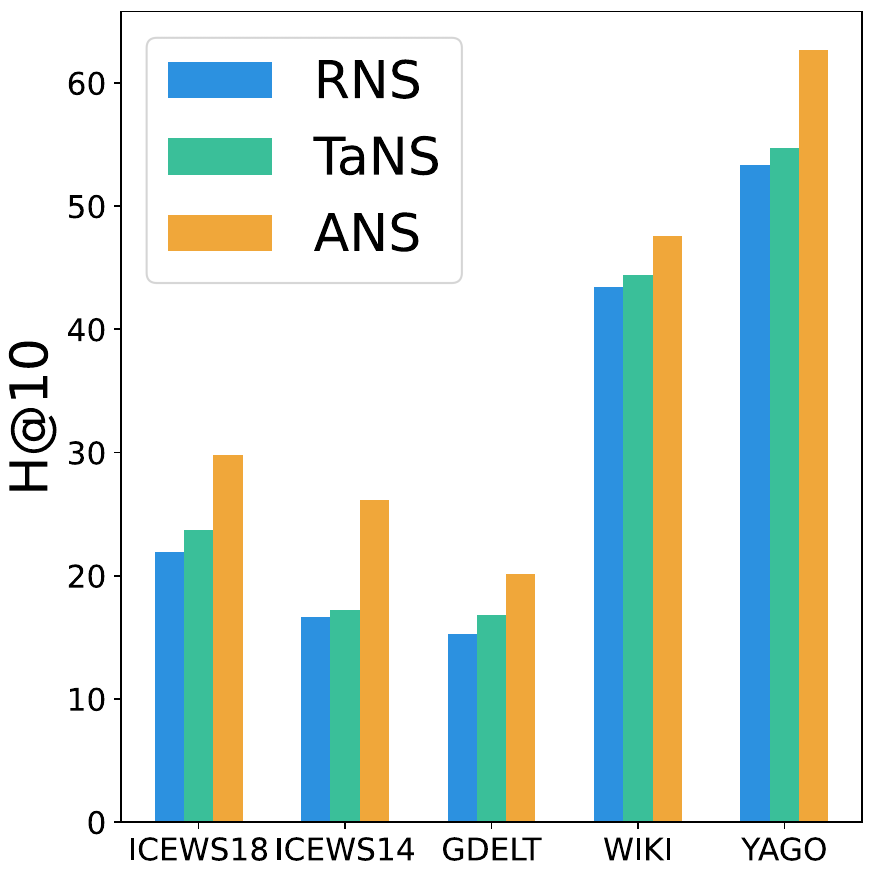}}
    \caption{Comparison results of different negative sampling methods on five datasets. RNS, TaNS and ANS represents random, time-aware and adversarial negative sampling, respectively.}
    \label{fig:9}
\end{figure}

For RQ2, we perform ablation analysis for the adversarial negative sampling method TKGAN. As shown in Table \ref{tab2}, we individually integrate the proposed TKGAN into five models from three types of baselines without the architectural design of TTT. The gain of TKGAN to the models is significant in almost all experimental settings compared to the traditional random negative sampling method. At the same time, only the individual metrics of individual models decreased, which is related to the choice of model framework and generator. Figure \ref{fig:9} compares TKGAN with random and time-aware negative sampling. The results suggest that TKGAN can bring better quality negative samples to the model and improve the model inference.

For the consideration of the TTT training structure in RQ3, in addition to the comparison of T3DM and TKGAN in Table \ref{tab2}, we apply it to more TKG models, as shown in Table \ref{tab3}. Due to their unique design, these models are not combined with negative samples, so we just use them to validate the TTT framework. From the results, it can be seen that for both CyGNet and Co-CENET, TTT obtains performance improvements in all metrics for all five datasets, and there are only two datasets with a decrease in H@10 and one H@1 on CENET. This phenomenon is mainly attributed to the limitations of CENET's unique contrastive learning mechanism, which results in the loss of distributional differences for model updating. The results demonstrate that our TTT training model and the designed auxiliary training tasks can improve the model's ability to cope with event distribution shift.

\subsection{Sensitivity Analysis (RQ4 \& RQ5)}
For RQ4, we compare different generators in TKGAN. As shown in Table \ref{tab4}, we choose TTransE, HyTE and DE-SimplE as generators, and discriminators are the same as in main experiments. From the results, we find that choosing TTransE as the generator gives the optimal performance with the most experimental setups (which is why choosing TTransE as the generator in main experiments). HyTE as the generator brings a better negative sampling effect for TATransE and HyTE, which is attributed to similar structural design and negative sampling process. The experiments show that for different TKG models, a model with similar structure can be chosen as the generator, and a suitable TKGAN design can improve their performance.

\begin{figure}
    \centering
    \subfloat{\includegraphics[width=0.12\textwidth]{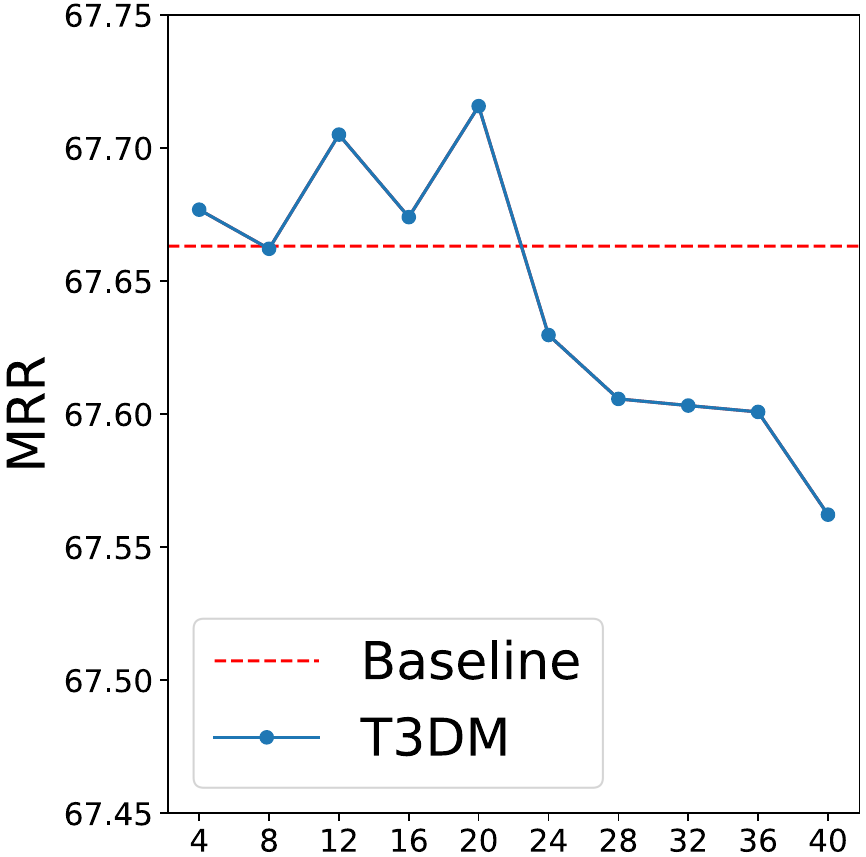}}
    \hfill
    \subfloat{\includegraphics[width=0.12\textwidth]{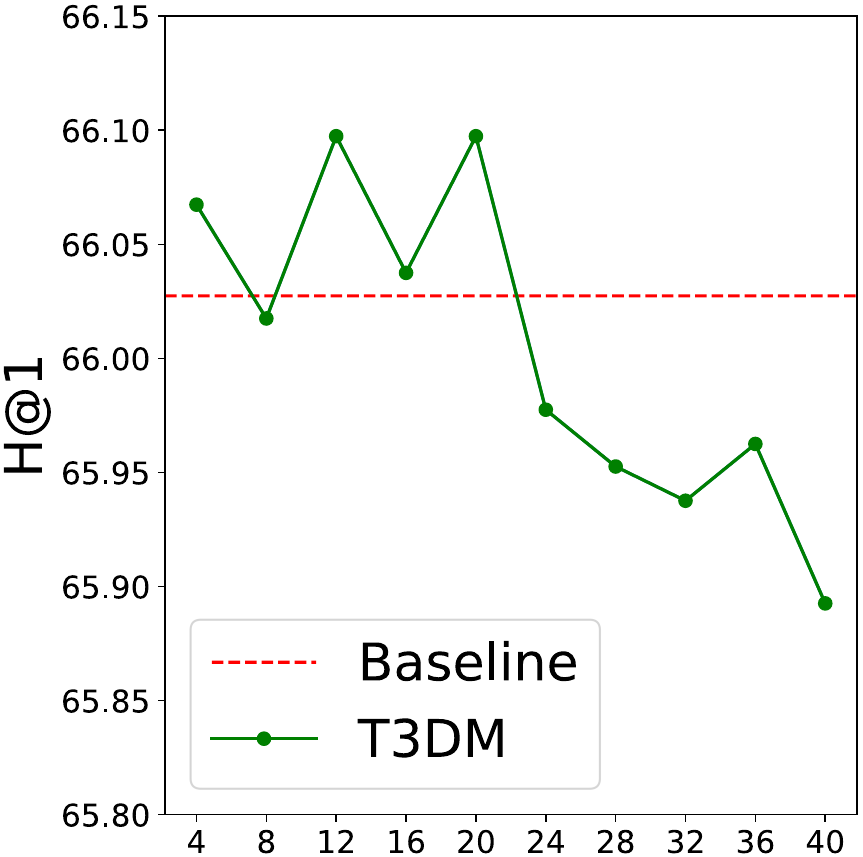}}
    \hfill
    \subfloat{\includegraphics[width=0.12\textwidth]{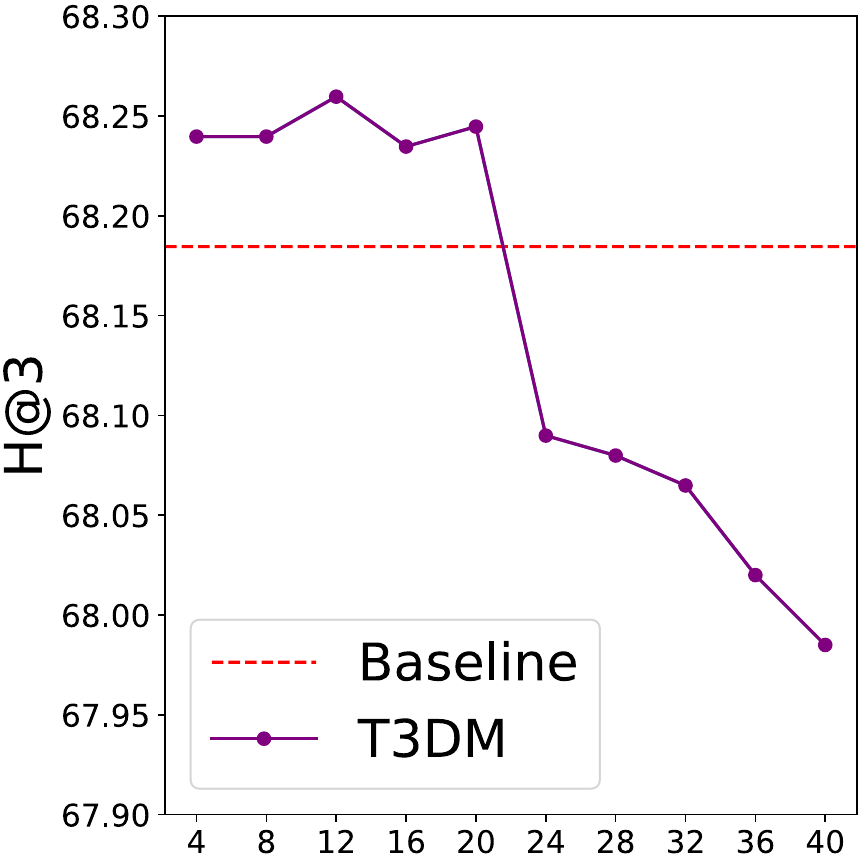}}
    \hfill
    \subfloat{\includegraphics[width=0.12\textwidth]{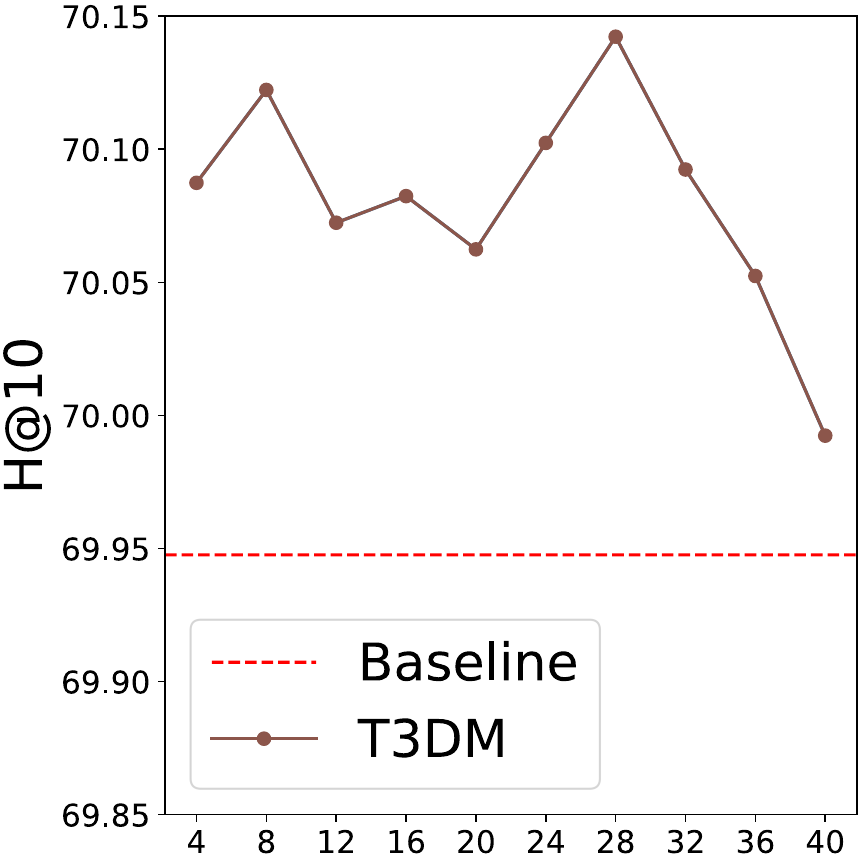}}
    \caption{Sensitivity analysis results of input sequence length hyperparameter in LSTM.}
    \label{fig:4}
\end{figure}

In addition, we analyse the hyperparameter of input sequence length in LSTM. As shown in Figure \ref{fig:4}, the model effect tends to be optimal when the length does not exceed 20 and decreases with increasing length. Nonetheless, the length variation can get a stable effect enhancement on H@10, which is further evidence of the significant gain of the TTT framework to the TKG models.

\begin{table}
\centering
\scriptsize
\setlength{\tabcolsep}{4pt}
    \renewcommand{\arraystretch}{0.95}
  \caption{Experimental results of different TKGAN group settings for TKG link prediction task on YAGO.}
\scriptsize
  \begin{tabular}{l|ll|ll|ll}
    \toprule
    Model G & \multicolumn{2}{l|}{TTransE}& \multicolumn{2}{l|}{HyTE}& \multicolumn{2}{l}{DE-SimplE}\\
    \midrule
    \multirow{2}{*}{Model D} & MRR & H@1& MRR & H@1& MRR & H@1\\
    ~ & H@3 & H@10& H@3 & H@10& H@3 & H@10\\
    \midrule
    \multirow{2}{*}{TTransE} &\textbf{41.84}&\textbf{27.05}&36.87&19.67&{40.21}&{25.17}\\
    ~&\textbf{54.47}&\textbf{62.65}&51.90&60.57&{52.69}&{61.38}\\
    
    \multirow{2}{*}{TATransE} &{54.37}&{47.92}&\textbf{54.98}&\textbf{49.03}&54.25&47.86\\
    ~&{59.31}&\textbf{65.27}&\textbf{59.65}&{65.12}&59.07&64.77\\
    
    \multirow{2}{*}{HyTE} &42.86&38.51&\textbf{56.32}&\textbf{50.65}&{53.85}&{47.91}\\
    ~ &46.47&52.39&\textbf{60.49}&\textbf{65.97}&{57.60}&{64.15}\\
    
    \multirow{2}{*}{TADistmult} &\textbf{62.82}&\textbf{59.37}&{51.79}&{44.50}&49.66&41.22\\
    ~ &\textbf{63.67}&\textbf{71.46}&{57.55}&{63.81}&56.42&63.38\\
    
    \multirow{2}{*}{DE-SimplE }&\textbf{30.80}&\textbf{34.39}&{22.65}&{14.90}&14.62&8.99\\
    ~&\textbf{38.69}&\textbf{47.15}&{24.74}&{38.63}&15.19&25.38\\
\bottomrule
\end{tabular}
	\label{tab4}
\end{table}

\section{Conclusions}
In this paper, we propose T3DM, a plug-and-play training method for distribution modelling in a TTT framework. Specifically, we are the first to introduce the TTT framework and the GAN model to TKGR. We use TTT framework to enhance the model's ability to model event distribution shift, and design an adversarial negative-sampling strategy to generate higher-quality negative quadruples. Extensive experiments on five datasets show that most baselines integrated with T3DM achieve better performance in link prediction.

\section*{Limitations}
This paper primarily focuses on TKGR, which aims to predict future unknown events based on existing knowledge. In designing the training tasks for the test phase, we only obtain labels through LSTM predictions of future distributions, which leaves room for improvement. In future research, we will attempt to design new self-supervised training tasks, such as integrating large language models to improve the encoding of entities and relations. Furthermore, we only evaluate the robustness of TKGAN with five baseline models. To more convincingly demonstrate the high-quality negative sampling of TKGAN, we will incorporate additional tasks and baseline models in future work.

\bibliography{ref}

\appendix

\section{Appendix}
\label{sec:appendix}

\subsection{Implementation Details}
The entire experimental implementation of T3DM is executed on a computational setup comprising an Intel (R) Core i9-10900K CPU and an NVIDIA GeForce RTX 3090 Ti GPU, based on the KGE open-source framework of PyTorch \cite{pytorch}. We focus on TTransE as the generator model and add the consideration of HyTE and DE-SimplE in the comparison. During training, the batch size of T3DM is set to 512, and the maximum training epoch limit is 1000. We choose Adam as the optimiser, and the learning rate is compared within a set of predefined values \{0.0001, 0.0005, 0.001, 0.01\} to select the optimal one. The input sequence length of LSTM is set to 20.

\subsection{Evaluation Metrics}
To evaluate the performance of TKGR models, we employ standard evaluation metrics like Mean Reciprocal Rank (MRR) and Hits@K. MRR represents the average of the inverse rankings of all samples:
\begin{equation}
\begin{split}
\text{MRR} = \frac{1}{2 \cdot \text{N}(\mathcal{G})} \sum_{o,s \in \mathcal{G}} ( \frac{1}{\text{RK}(o_{t} | o_{p})} + \frac{1}{\text{RK}(s_{t} | s_{p})} ),
\end{split}
\end{equation}
where, $\mathcal{G}$ denotes the quadruple set, $\text{N}(\cdot)$ denotes the number of elements, $o_{t}$ and $s_{t}$ denote the true entities, while $o_{p}$ and $s_{p}$ denote the predicted entities. $\text{RK}(\cdot)$ denotes the recommendation ranking of the correct answer. Hits@K (where K=1,3,10) represents the proportion of test samples which are ranked in top K positions:
\begin{equation}
\begin{split}
\text{Hits@K} = \frac{1}{2 \cdot \text{N}(\mathcal{G})} \sum_{o,s \in \mathcal{G}} ( \textbf{1}\{\text{RK}(o_{t} | o_{p}) \leq K\}
\\ + \textbf{1}\{\text{RK}(s_{t} | s_{p}) \leq K\}).
\end{split}
\end{equation}

\subsection{Statistics of Datasets}
Five public datasets differ in their factual representations: facts in GDELT and ICEWS are based on a specific time point, while facts in YAGO and Wikidata are based on time intervals. Statistics details of five datasets are shown in Table \ref{tab1}.

\begin{table}
\scriptsize
\setlength{\tabcolsep}{1.7pt}
    \renewcommand{\arraystretch}{0.95}
  \caption{Statistics details of five publicly available datasets.}
  \begin{tabular}{llllllll}
    \toprule
    Dataset	& Entities & Relation &Time & Training & Validation & Test&Interval\\
    \midrule
    ICEWS14	& 12,498 & 260 &365& 323,895 & - & 341,409&24 hours\\
    ICEWS18	& 23,033 & 256 &304& 373,018 & 45,995 & 49,545&24 hours\\
    GDELT & 7,691	& 240 &2,751& 1,734,399 & 238,765 & 305,241&15 mins\\
    WIKI & 12,554 & 24 &232& 539,286 & 67,538 & 63,110&1 year\\
    YAGO & 10,623 & 10 &189& 161,540 & 19,523 & 20,026&1 year\\
  \bottomrule
\end{tabular}
	\label{tab1}
\end{table}

\subsection{``Bern'' sampling implementation details.}
We set different probabilities for replacing the head or tail when corrupting the quadruples, which depends on the mapping property of relation. We tend to give more chance to replacing the head if the relation is 1-to-N and replace the tail if N-to-1. In this way, the chance of generating false negative labels is reduced. Specifically, the average number of tail per head is denoted as $Nt$, and the average number of head per tail is denoted as $Nh$. We corrupt the quadruple by replacing head with probability $\textstyle\frac{Nt}{Nt+Nh}$, and replacing tail with probability $\textstyle\frac{Nh}{Nt+Nh}$.

\end{document}